\documentclass[manuscript,screen,review = false]{acmart}

\AtBeginDocument{%
  \providecommand\BibTeX{{%
    \normalfont B\kern-0.5em{\scshape i\kern-0.25em b}\kern-0.8em\TeX}}}

\usepackage{array}

\begin{document}

\title[ABM to Simulate the Expansion of a Virus]{Agent-Based Model: Simulating a Virus Expansion Based on the Acceptance of Containment Measures}

\author{Alejandro Rodríguez-Arias}
\email{alejandro.rodriguez.arias@udc.es}
\orcid{0000-0002-0140-7473}
\affiliation{%
  \institution{Research Center on Information and Communication Technologies}
  \streetaddress{Campus de Elviña s/n}
  \city{A Coruña}
  \country{Spain}
  \postcode{15071}
}

\author{Amparo Alonso-Betanzos}
\affiliation{%
  \institution{Research Center on Information and Communication Technologies}
  \streetaddress{Campus de Elviña s/n}
  \city{A Coruña}
  \country{Spain}
  \postcode{15071}
}

\author{Bertha Guijarro-Berdiñas}
\affiliation{%
  \institution{Research Center on Information and Communication Technologies}
  \streetaddress{Campus de Elviña s/n}
  \city{A Coruña}
  \country{Spain}
  \postcode{15071}
}

\author{Noelia Sánchez-Maroño}
\affiliation{%
  \institution{Research Center on Information and Communication Technologies}
  \streetaddress{Campus de Elviña s/n}
  \city{A Coruña}
  \country{Spain}
  \postcode{15071}
}

\renewcommand{\shortauthors}{Alejandro, et al.}

\begin{abstract}
Compartmental epidemiological models categorize individuals based on their disease status, such as the SEIRD model (Susceptible-Exposed-Infected-Recovered-Dead). These models determine the parameters that influence the magnitude of an outbreak, such as contagion and recovery rates. However, they don't account for individual characteristics or population actions, which are crucial for assessing mitigation strategies like mask usage in COVID-19 or condom distribution in HIV. Additionally, studies highlight the role of citizen solidarity, interpersonal trust, and government credibility in explaining differences in contagion rates between countries.
Agent-Based Modeling (ABM) offers a valuable approach
to study complex systems by simulating individual components, their actions, and interactions within an environment. ABM provides a useful tool for analyzing social phenomena. In this study, we propose an ABM architecture that combines an adapted SEIRD model with a decision-making model for citizens. In this paper, we propose an ABM architecture that allows us to analyze the evolution of virus infections in a society based on two components: 1) an adaptation of the SEIRD model and 2) a decision-making model for citizens. In this way, the evolution of infections is affected, in addition to the spread of the virus itself, by individual behavior when accepting or rejecting public health measures. We illustrate the designed model by examining the progression of SARS-CoV-2 infections in A Coruña, Spain. 

This approach makes it possible to analyze the effect of the individual actions of citizens during an epidemic on the spread of the virus.
\end{abstract}

\ccsdesc[500]{Computing methodologies~Model development and analysis}

\keywords{Agent-based model, SEIRD model, COVID19, virus spread}
%

\maketitle

\section{Introduction}
Predicting the evolution of a phenomenon or explaining patterns of behavior observed in the past requires an in-depth analysis of the incident of interest. Usually, using real systems for this purpose is problematic because it can be expensive, time-consuming, impossible to recreate in a real environment, or ethically questionable. For example, in the case of a virus, it is not acceptable to test its ability to spread in an actual situation, neither in humans nor in animals.

The current pandemic caused by SARS-CoV-2 has shown that not even the first world powers have highly effective systems for epidemiological control. Thus having reliable, fast, and easy-to-adapt tools that would allow simulating a large number of different situations would be convenient to use as an aid in decision-making for the control of several epidemics. 

As a solution to this problem, models can be used, understood as a simple but useful representation of the system to be studied and that seeks to answer a specific research question. There are numerous ways to define models from different areas of research, but Agent-Based Modeling (ABM) has established itself as a common methodology \cite{niazi2011agent}. Agent-based models represent systems by simulating the actions of their individual components (agents, which act autonomously), and the interaction between them and their environment, allowing complex processes and systems to be solved.  These autonomous agents are placed in some virtual environment  -it could be a geographical representation or a network of agents-, and they are capable of perceiving changes in their environment, interacting with it and with other agents in the model, and adapting their behavior if necessary, thus allowing for solving complex problems. This type of modeling can be used to represent practically any phenomenon. 

A key feature of ABM is that it allows us to tackle or study problems related to the concept of \textit{emergent phenomena} \cite{railsback2019agent}. This means examining system dynamics that appear from the behavior and interaction of the individuals that make up the system, thus allowing the study of questions related to how the global behavior of society links to individual behavior. An example could be the behavior of ants when searching for food, that can be divided into two states: (1) it has food and heads towards the hive expelling pheromones along the way,  and (2) it has no food, so it looks for a nearby pheromone trail, if it exists, or it moves erratically until it finds food. For a pheromone trail to be powerful enough to guide an ant, many other ants must discover the food source on their own. The emerging phenomenon is that the hive learns to make a systematic and optimal collection of all food sources from the results of each individual behavior.


A classic and standard way of modeling the spread of a virus in a society is through the compartmental model of epidemiology \textit{susceptible-exposed-infected-recovered-dead} (SEIRD). The SEIRD is a mathematical model that consists of five compartments or states for the individuals of a population at risk of infection: \textit{susceptible} (S) the status for individuals susceptible to being infected, \textit{exposed} (E) the status for individuals that are incubating the virus but cannot transmit it to others yet, 
\textit{infectious} (I) the status for individuals who can infect others, 
\textit{recovered} (R) the status for individuals already cured of the virus,
and \textit{dead} (D) the status for individuals who have died from the virus.

In the study of a virus spread, in addition to the evolution that is extracted from the basic epidemiological model, it is of vital importance to predict the behavior of individuals to face possible policy measures adopted to mitigate this viral expansion. The main problem of the SEIRD model, as well as any other classical epidemiological model, is that the only information that defines an individual is that related to their epidemiological status, and so all those who are in the same state are considered identical. A solution to this problem is integrating the SEIRD in an ABM, allowing each agent in the model to represent a single individual with its characteristics, both in terms of its sociodemographic variables and its epidemic state. The psychological needs of each individual and the importance it assigns to them determine the compliance or not with the measures imposed. In this way, the SEIRD model can include individual properties that may affect viral expansion, from individual actions to specific genetic traces.

In this paper, we propose an agent-based model architecture that can be easily adapted to any virus or epidemic that meets the SEIRD epidemiological transition model. To represent individuals, the ABM will use an adaptation of the HUMAT \cite{Antosz2019} decision-making model, designed in the SMARTEES project \footnote{The SMARTEES Project Local Social Innovation: \url{https://local-social-innovation.eu/}} under the European Horizon 2020 program, in which an ABM was designed to study the citizens’ acceptance of different social innovations considering individual psychological needs. This model, developed in collaboration with psychologists and sociologists, will serve as the basis for this work with the relevant adaptations to simulate society's acceptability of preventive measures and their effect on the expansion of the virus. Using HUMAT, the ABM allows to load in its architecture the socio-demographic data of the population to be represented in order to evaluate the acceptance by the society of the preventive measures against the simulated virus. This decision making will depend on the needs of each individual, so to feed the model, in addition to the actual socio-demographic data, it will be necessary to obtain the needs that define the character of each individual. The proposed model will make it possible to analyze the evolution of infections in a human society based on the characteristics of the virus and the individual decisions of each citizen.

The document is structured as follows. Section \ref{sec-stateart} describes the state of the art in social and epidemiological problems. Section \ref{sec-Agentes} explains the adaptation of the HUMAT decision-making model to the agent system for simulating the expansion of a virus. Section \ref{sec-criticalnodes} defines the adaptation of certain relevant actors (like policymakers) to the model and its influence on HUMAT agents. The adjustment of the SEIRD to the model is described in section \ref{sec-SEIRD}. Section \ref{sec-representacionsociedad} specifies the representation and mobility of the society in the model. Section \ref{sec-interfaz} describes the interface of the model. While section \ref{sec-loop} explains the model execution loop. Section \ref{sec-Ejemplo} presents an example of adapting the model to the case of SARS-CoV-2 in A Coruña (a city in the north of Spain) and Section \ref{sec-Escenarios} includes possible scenarios and results for the example case. Finally, the last section ends with the exposition of the conclusions.

\section{State of the art}\label{sec-stateart}
Combating pandemics of emergent and re-emergent infectious diseases has been broadly studied by scientists from different fields. However,  the problem stems mainly from two reasons: (1) the continuous and ever-lasting mutations of the viruses, and (2) the complexity in the disease transmission mechanism \cite{siettos2013mathematical}. Efforts have focused on predicting, assessing, and controlling potential outbreaks \cite{siettos2013mathematical}. The pandemic caused by COVID-19 has led to a notable increase in these efforts, so the bibliography found to deal with this virus is very extensive and is growing every day. Similarly, there are many mathematical models available to analyze the epidemiology events, covering statistical-based methods, state-space models (such as SEIRD and ABM), or even machine learning models\cite{siettos2013mathematical,hernandezpereira2022}.  Therefore, this state of the art cannot be exhaustive and will focus on remarkable publications using similar techniques to the one proposed inhere i.e., employ either a SEIRD, an ABM representing a SEIRD or a different aspect of the pandemic, or a combination of both.

On the first hand, many models have appeared  with classical approximations of the SIR or SEIRD models for the COVID-19 pandemic \footnote{At the time of writing, the number of references provided for Google Scholar when looking for COVID and SEIRD is over 19500. This number is much greater if we look for COVID and SIR}, thus for the sake of brevity, we will include just a few.   Loli Piccolomini \& Zama proposed a forced SEIRD differential model for the analysis and forecast of the COVID-19 spread in Italian regions, using data from the Italian Civil Protection Department \cite{loli2020monitoring}. Ala'raj \emph{et al.} combined two different mathematical models (SEIRD and ARIMA, Autoregressive Integrated Moving Average) to study the properties associated with this pandemic \cite{ala2021modeling}. Menda \emph{et al.} used, in addition to the SEIRD, a neural network to predict the infection rate and fit a single model to data from multiple counties in the United States exhibiting different behavior \cite{menda2021explaining}. However, as already mentioned, the SEIRD model is not sufficient when we want to observe not only the aseptic transmission of the virus, but also how the behavior of individuals influences its spread.

On the other hand, the capability of the ABM to represent societies, through the characterization of individuals, allows capturing directly and effectively the expansion of a virus, whether in an animal or human population. For this reason, it is becoming a standard in the simulation of viral spreads and has been used to represent, among others, the spread of viruses such as the H5N1 virus (avian flu)\cite{abm2}, H1N1 virus (swine flu) \cite{abm4}, tuberculosis, \cite{abm3} respiratory viruses\cite{abm6}, or SARS-COV-2 \cite{lorig2021agent, tang2022agent}.  Silva \emph{et al.} \cite{silva2020covid} proposed a SEIR agent-based model that aims to simulate the pandemic dynamics using a society of agents emulating people, business, and government. In that way, they assess the economic effects of seven different scenarios with specific social-distancing interventions (from doing nothing to lockdown). Kerr \emph{et al.} \cite{kerr2021covasim} developed an open-source model to help project epidemic trends, explore intervention scenarios, and estimate resource needs, with the aim of being capable of informing real-world policy decisions. Hinch \emph{et al.} \cite{hinch2021openabm} presented an agent-based simulation of the Covid epidemic including detailed age-stratification and realistic social networks, which can evaluate non-pharmaceutical intervention (NPI), including both manual and digital contact tracing, and vaccination programmes. Badham \emph{et al.} \cite{badham2021justified}  used the SEIRD model in conjunction with an ABM that allows representing the individual characteristics of each citizen and permits political decision-making by simulating the impact of different preventive measures, such as social distance. However, the common drawback of these works, is that agents are unable to make decisions about their own behavior in these models, i.e., these models do not take into account the human psychological factor, and assume that all citizens abide by a rule once established, although reality has shown us that this is not the case, either consciously (seeking social contact) or unconsciously (inappropriate use of a mask).

The behaviour of citizens is well defined by Palomo  \emph{et al.} \cite{palomo2022agent}, as they proposed an ABM with two well-differentiated submodels, the first one is in charge of the epidemiological evolution through a SEIRD, and the second one defines citizens' behavior. These citizens have their displacement agenda in the virtual environment,  make decisions about which preventive measures they are going to comply with, and their citizen profile is defined by the geographical area in which they are located. One of the main problems with this model is that it ignores social relations, meaning that they do not consider how they affect the evolution of individual needs, and, consequently, how they influence their behaviour. Moreover, a recently published macro-study \cite{TheLancet} discovers the significant importance of citizen solidarity, interpersonal trust and government credibility in explaining the differences in contagion between countries and rules out medical and technical aspects. So all these essential aspects need to be represented in the model.

\section{Citizen Agents}\label{sec-Agentes}

The are several crucial components of the proposed ABM: the agents and their environment, the interactions between them, the decisions they make, and the influence of all these on the spread of the virus. In this section we will focus on the main component that is the agents. In ABM, an intelligent agent of the model is considered to be a computational system located in an environment, which can perceive changes in that environment and has the ability to react appropriately and autonomously to these changes.

The agents used in the proposed model represent citizens. First, the agents are individualized by using the sociodemographic characteristics shown in table \ref{tabla-agentes} which were chosen to model the expansion of SARS-CoV-2, as it will be explained later. Moreover, according to the psychologists who participated in the SMARTEES project, an important aspect of individualizing agents is the representation of their needs, such as their need for well-being. It is worth highlighting the need of belongingness to the group, which defines the agent's need to be similar to other members of his group and which will have a special weight in actions that the agent performs.
Each citizen seeks to satisfy these needs and to do so, to a greater or lesser extent, influences his acquaintances in the same way that they also influence him, which conditions his decision making.

One of the factors determined as key for reducing the impact of the pandemic has been individual responsibility. Therefore, in the proposed model, the decision-making of the agent (citizen) consists of complying (or not) with the preventive measures imposed by the authorities to limit the spread of the virus. To carry out these decisions, these agents are developed from an adaptation of the HUMAT decision-making model which will be detailed in the next section.

\begin{table*}
\caption{Variables that characterize the model agents}
\begin{center}
\begin{tabular}{m{6cm}m{8cm}}
\toprule
\textbf{Parameter} & \textbf{Value} \\
\midrule
gender & man/woman\\
age & integer in [18,100]\\
family & one-person household/ father or mother alone with children / father or mother alone with children and other people/ Couple with children/ couple with children and other people/ Couple without children / Another type of home \\ 
cotagge & yes/no\\
economic activity & employee/ unemployed/ autonomous/ civil servant / executive or director/ college student / retired\\
essential worker & yes/no\\
net monthly salary in euros & no income / less than 1000 / [1000-1500]/  [1501-3000]/  [3001-4500]/  [4501-6000]/ more than 6000 \\
census tract & census tract code\\
\bottomrule
\end{tabular}
\label{tabla-agentes}
\end{center}
\end{table*}

\subsection{Decision-making model: HUMAT}\label{sub-HUMAT}

HUMAT \cite{Antosz2019} is an architecture designed for the construction of artificial populations, where each agent makes its own decisions based on its needs and the influence of its social network. In this model, decision making, social interactions, and agent cognition are based on social science theories.

\begin{figure}
\centerline{\includegraphics[width=0.85\textwidth]{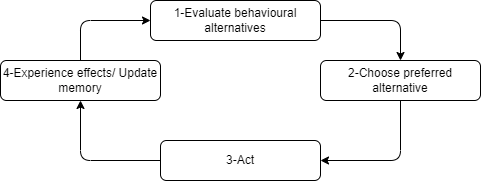}}
\caption{HUMAT decision-making cycle}
\label{fig-humat}
\end{figure}

The agent's decision-making process is divided into four phases that are constantly repeated during the model's life cycle, as can be seen in figure \ref{fig-humat}. Each of these phases will be discussed in detail below.

\subsection{Phase 1: Evaluate behavioural alternatives}

The team of psychologists and sociologists - who collaborated in creating the HUMAT model- defined at least two categories of needs that are always present in individuals in their decision-making. The first one is related to hedonism, i.e., the individual's need for pleasure, for example, health or comfort. The second one is directly related to the agent's social network and it is the need for belongingness, which is the human emotional need to be an accepted member of a group. More needs can be added depending on the problem to be modeled.

Each citizen values the importance of each of these needs differently in his/her decision-making. Depending on the importance that each individual gives to these needs and how he/she can satisfy them according to the circumstances surrounding him/her, the individual will reach a certain \emph{level of satisfaction} for each need. In the model, the value of the \emph{importance} given to each need is in the range [0,1] and the value of \emph{satisfaction} of each need will be in the range [-1,1]. The initial values of satisfaction and importance must be initialized using an external data source such as data obtained from a survey.


The evaluation $E$ of the satisfaction $S$ of a need $n$ for a given behavioural alternative $b$ depends on the importance $I$ given by the agent and is calculated according to the equation \eqref{eq-needevaluation}
\begin{equation}
        E_{b,n}= S_{b,n} * I_{n} \quad\quad n = 1 \ldots N , \quad b = 1, 2 \label{eq-needevaluation}
\end{equation}

Those evaluations that result in a positive value $(E^{+})$ will be considered ``satisfying"  while those that have a negative value $(E^{-})$ will be considered ``dissatisfying". Evaluations with a value of 0 will be considered ``neutral".

A citizen can choose between two behavioral alternatives: accept or reject preventive measures. This decision will be made based on the overall satisfaction of the needs that each alternative is expected to produce. The overall expected satisfaction $O$ for a behavioural alternative $b$ is given by the equation \eqref{eq-overall}. 

\begin{equation}
        O_{b}=\frac{\sum_{n=1}^{N} E_{b,n}}{N} \label{eq-overall}
\end{equation}

When one of the behavioral alternatives evokes sufficient levels of satisfaction in one or several groups of needs but also certain levels of displeasure in others, a \emph{cognitive dissonance} appears. Thus, the level of dissonance generated by a behavioral alternative \textit{b} (\(D_{b}\)) is given by equation \eqref{eqdissonance}

    \begin{equation}
        D_{b}=\frac{2d_{b}}{d_{b} + c_{b}} \label{eqdissonance}
    \end{equation}
\noindent where:

    $$ d_{b} = min \left(\left|\sum_{n=1}^{N}  E_{b,n}^{+}\right|, \left|\sum_{n=1}^{N}   E_{b,n}^{-}\right| \right) $$
    $$ c_{b}  = max \left(\left|\sum_{n=1}^{N}  E_{b,n}^{+}\right|,  \left|\sum_{n=1}^{N}   E_{b,n}^{-}\right|\right) $$

If this dissonance exceeds a tolerance threshold, then the agent is faced with a \emph{dilemma}. A given need generates a dilemma if its evaluation is positive ($E^{+}$) where all the others are negative ($E^{-}$) or vice versa. This means each of the other evaluations individually must be of the opposite sign to that of the given need. Therefore, there could be as many dilemmas as needs. Later we will see which strategies the agent uses to solve the dilemmas.
    
\subsection{Phase 2: Choose preferred alternative}

Once the citizen obtains the expected satisfaction for each of his behavior alternatives, he chooses the one that produces higher levels of satisfaction. However, if both behavioral alternatives are similar --less than 10\% of the range as a difference--, the agent explores in depth the difference between both alternatives. Scanning happens in the following order:

\begin{enumerate}
  \item Exploration of the cognitive dissonance (see equation \ref{eqdissonance}) for each of the alternatives. If this dissonance is different enough (more than 10\% of the range between both dissonances), the alternative that generates less dissonance is selected. Otherwise, the exploration continues.
  \item When faced with a difficult decision, an agent tends toward hedonism, so the hedonistic need is explored. If the difference between the evaluation of the need for well-being between both alternatives is significant, the one with the highest evaluation will be selected. 
  \item Finally, if none of the above criteria is applicable, the agent randomly selects an alternative.
\end{enumerate}

\subsection{Phase 3: Act}

The actions available to agents consist of different forms of communication with members of their social networks.  When the selected behavioral alternative generates a dilemma agents act to reduce their dissonance and will try to either persuade or seek advice on the benefits of the selected behavioral alternative. Therefore, there are two main types of actions: \textit{Inquiring} and \textit{Signaling}. Inquiring involves questioning other agents about their opinion of the preventive measures and its is employed when the agent has non-belongingness dilemma. Signaling consists of informing other agents about one's own opinion on the measures, and is used when the agent has a belongingness dilemma. 

In order to simulate as realistic an environment as possible, also from time to time, citizens have random conversations similar to \textit{Signaling} even if they do not experience any dilemmas.

Every time a communication ends, the agents involved reassess their dissonances to see if they have been reduced.

The way to select the agent, on whom the action of inquiring or persuading is going to be carried out, depends on several factors:

 \begin{enumerate}
  \item Existing relationship. The agent to interact with must belong to the same social network.
  \item The current behavior of the agent. In the case of \textit{Inquiring}, an agent with the same behavior is preferred, while for a \textit{Signaling} action an agent with the opposite behavior is preferred.
  \item The persuasiveness of the agent sending the message on the receiving agent, that depends on how difficult it was to persuade the agent in the past.
\end{enumerate}

\subsection{Phase 4: Experience the effects and update memory}

The ability of an agent to influence another varies depending on the agents involved in the action, as each citizen has a specific trust value for each agent that is part of his network of influence. However, an agent does not radically change its mind depending on the influence exerted by the agents around. In fact, its own opinion prevails (taking a value that goes from 60 \% to 100 \%) while the opinion of others can only cover the range [0 \%-40 \%].

The actual influence oscillates in this range based on two factors:

\begin{itemize}

  \item The trust of the agent that is influenced in the agent who exerts the influence. This trust does not have to be reciprocal between both agents.
  \item The similarity of needs between both citizens, that is, the importance given by them to the different needs is similar. If the evaluations of the need (see equation \eqref{eq-needevaluation}) of both agents have a different sign, the value of the \emph{similarity of needs} is 0. In another case, this similarity  is calculated as
\begin{equation}
        M_{b,n}= 1 - |I_{e,b,n} - I_{o,b,n}| \label{eq-needsimilarity}
\end{equation}

 \end{itemize}
 
For a maximum influence of 40\% to occur in communication, it is necessary that the needs of both agents are identical and that the trust of the agent receiving the communication in the agent that carries it out is maximum. On the other hand, if the needs are completely different or there is not even the slightest trust, communication will have a minimum influence of 0\%. The final \emph{persuasion} of the agent that is trying to influence is calculated as

\begin{equation}
        P_{b,n}= \alpha * T * M_{b,n} \label{eq-persuasion}
\end{equation}

\noindent where \emph{T} is the trust of the influenced agent on the influencer and $\alpha \in [0,0.5]$ is an arbitrary number that ensures that the agent's previous opinion weighs more in its final satisfaction than the opinion received during the communication \footnote{0.4 has been selected after a calibration process for this model \cite{bouman2021report}}.

The new satisfaction \textit{N} value is calculated using the equation \eqref{eq-newsatisfaction}. Where \textit{b} is the behavioural alternative, \textit{n} is the need, \textit{S} the need satisfaction, \textit{I} the importance\textit{o} the influencing agent, and \textit{e} the influenced agent.

\begin{equation}
        N_{b,n}= (1- P_{b,n}) * S_{b,n,e} + P_{b,n} * S_{b,n,o}  \label{eq-newsatisfaction}
\end{equation}

\noindent where \emph{S} is the previous satisfaction of the need.

Once an action ends, the dissonances are re-evaluated with the updated satisfaction values.

\section{Critical Nodes}\label{sec-criticalnodes}
The model may also include some critical nodes, those are agents representing groups or institutions that may have a relevant role in the evolution of the acceptability of preventive measures by citizens, such as the city council, the press or other media, or the political opposition. These agents have motivations that do not have to be aimed at reducing the effect of the pandemic.

A critical node has its own social network, and only the agents in it can be influenced by a direct communication from the critical node. It is generated as a random network during model initialization and its size is defined in the configuration file of each critical node. On the other hand, citizens have a certain trust value in each critical node that determines the strength of the influence that node can exert.


Each critical node has its strategy aimed at achieving its own goals. The only action these nodes can carry out is to communicate with citizens, so these strategies are reduced to attempts to influence the opinion or behavior of citizens. The communications carried out by the critical nodes are defined by the following parameters:

 \begin{itemize}
  \item Main critical node: the node that initiates the communication.
  \item Orientation of communication: communication can be aimed at promoting compliance with the preventive measures or at showing doubts about their effectiveness and therefore reducing their use.
  \item \emph{Dates}  when communications start and end and the \emph{frequency} of the messages during that period.
  \item Reach: the percentage of the critical node's social network which is reached by this communication.
  \item Secondary critical node: the node that performs the communication designed by the primary critical node. As each node's network is different, specific nodes may be interested in using the services of another node to increase the range of communications. For example, a denialist association may be interested in using the press to broaden the reach of its messages.
\end{itemize}

Each critical node has a predefined communication plan that tells it when to send a message to a certain part of its social network. This message acts similarly to the agents' communication (see \eqref{eq-newsatisfaction}. The \textit{P} persuasion of the agents upon receiving a communication from a critical node is calculated following equation \eqref{eq-persuasioncriticalnode}.

\begin{equation}
        P =  T * 0.2  \label{eq-persuasioncriticalnode}
\end{equation}

\section{Epidemiological model}\label{sec-SEIRD}

Besides the ABM, a second system in the model is responsible for simulating the spread of the virus in the modeled society, for which we use an adaptation of the standard SEIRD epidemiological compartmentalized model \cite{ma2009mathematical}. Normally, in a compartmentalized model, a population is divided into closed groups and all individuals belonging to a group are assumed to share the same characteristics as if they were clones. In the adaptation proposed in this work, the individuals (the agents) that form a population are treated as unique individuals with their own characteristics and needs.

In the SEIRD model, individuals go through a series of epidemiological transitions between the following states: \textit{susceptible} to the disease (S), \textit{exposed} or incubation of the virus (E), \textit{infectious} (I) and the two final states that are \textit{recovered} (R) or \textit{dead} (D). For this work, an adaptation of the standard SEIRD model will be used, dividing the infectious state into 3 substates that allow evaluating hospital pressure: a) \textit{infectious} in the community, b) \textit{hospitalized} and c) in intensive care unit (ICU).

\begin{figure*}
\centerline{\includegraphics[width=10cm]{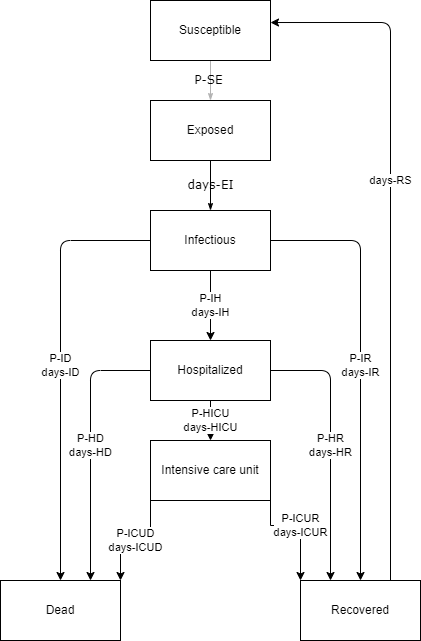}}
\caption{Transition diagram between epidemic states. Boxes represent states and arrows represent valid transitions between states. \textit{P-XY} is the probability that the transition from state \textit{X} to \textit{Y} occurs; \textit{days-XY} is the number of cycles that must elapse for the state change to be effective.}
\label{fig-TransiciónEntreEstados}
\end{figure*}

In figure \ref{fig-TransiciónEntreEstados} we can see the transition diagram between states of the epidemic. The transitions between states are assigned a probability of change and a number of days for the transition to be effective. Some of the transitions lack probabilities as they are mandatory transitions. For example, once one reaches the recovered state and gains some resistance to the virus, this resistance wears off over time and a mandatory transition to the Susceptible state is made. The values of the variables affecting the transitions are set through a configuration file that allows the model to be adapted to any virus that complies with the described state diagram.

During each cycle of the simulation, the state of the model agents will be evaluated and a transition between states of the epidemiological diagram will be executed, if necessary.

\section{Representation of the society}\label{sec-representacionsociedad}

An ABM must also represent the environment in which the agents are situated and simulate their relationships with their environment. In this work this involves defining the geographical environment, the movement of agents in this environment, and the social networks in which the agents are related.

\subsection{Virtual environment}\label{sub-entornovirtual}

\begin{figure}
\centerline{\includegraphics[width=7cm]{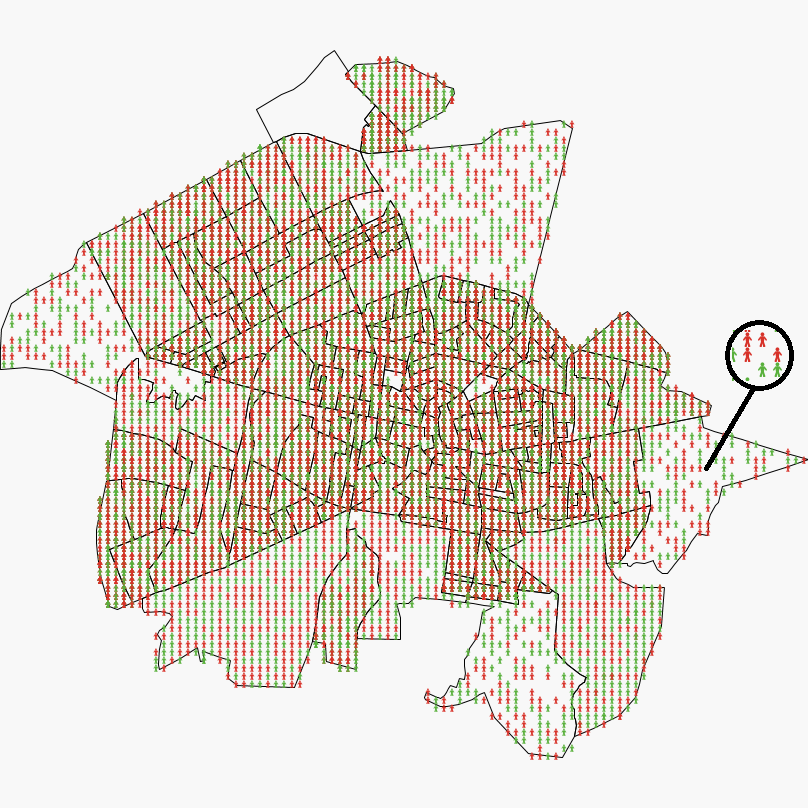}}
\caption{Representation of a geographical environment: a map of a city divided into census tracts. Agents are represented in green (healthy) or red (infected).}
\label{fig-city}
\end{figure}

First, the physical environment in which the agents coexist must be recreated. Concerning the geographical environment, a two dimensional board of 50x50 cells is used to represent the city through its census tracts. Some of these cells will be associated with places of work, study, essential purchases, or leisure, that is, places of a given relevance in the spread of a virus. It should be noted that each cell represents a large area of the city, so the same cell may contain more than one of these places. Each agent will be located in a single cell and represented by a human silhouette. In Figure \ref{fig-city}, it can be seen an example of this representation in a map divided into sections where the agents are placed, represented in green (healthy) or red (infected). The model can load one city or another by changing the shapefile with the census tracts data.  The representation of the city is done by its division in census tracts to facilitate the generation of a distribution that coincides with a real one, both in general demographic characteristics and in its spatial distribution. This spatial distribution is relevant for generating social networks.

Secondly, social networks allow us to unite different agents through links and become essential to facilitating agents to interact with each other. These links are directed and weighted, where weights represent the trust that the source agent has in the destination agent. This trust affects their communication (see section \ref{sec-Agentes}).

In the model designed, two social networks have been considered: 1) the network of neighbors and 2) the network of friends. A proximity relationship allows us to form the first social network, that is,  each agent establishes connections with the agents who live nearby. We have used the concept of social circles \cite{hamill2009social}, which allows for creating a network based on the current distance on the board, ensuring that it is reciprocal between the agents that belong to the network - i.e., if agent A is a neighbor of agent B, then agent B is a neighbor of agent A. These networks are limited in size, vary its size between individuals, have a high clustering capacity, and naturally recreate a network of neighbors. The configurable parameter \textit{social-reach} indicates the reach (in board cells) that each of these social circles will have.

Although networks based on social circles correctly represent neighbors' networks, they fail to adequately capture the friends' networks, since it would not be realistic that an entire network of contacts of an agent was living in its cell.

To solve this problem, the model has a second social network: the network of friends,  generated as a random network with two restrictions: 1) there will be a minimum number of links per agent (configurable parameter \textit{numfriends}) and 2 ) these links will be formed based on a homophily relationship concerning the age attribute, i.e., both agents are in a similar age range. In our case, the homophily condition intends to prevent networks of friends with very different age ranges, for example, an 18-year-old agent in a network of friends made up of agents 60 years of age or older.  However, to make the model more realistic, agents may have relationships with others who are far from their age range with a certain probability \textit{random-friend}. Thus, it allows to form networks of friends with individuals of very different ages but prevents this from being the norm.

Table \ref{tabla-socialnetworks} summarizes the configurable parameters used for the creation of social networks.

\begin{table*}
  \caption{Social network configuration parameters}
  \label{tabla-socialnetworks}
  \begin{tabular}{ccc}
    \toprule
    \textbf{Parameter} & \textbf{Use} & \textbf{Default values} \\
    \midrule
    social-reach & Agent social circle radius & 1\\
    num-friends & Minimum number of agents in the social network of friends & 5\\
    random-friend & Probability of establishing a friendship with an agent who is not of a similar age. & 0.05\\
    \bottomrule
  \end{tabular}
\end{table*}

\subsection{Mobility of citizens}

As described in section \ref{sub-entornovirtual}, the world proposed in this work consists of a map of 50x50 cells to represent the city in census tracts. Bearing in mind that these cells cover a large enough geographical area, there may be different locations in each one, such as places representing home, leisure, or work. Each agent recreated in the model has a variable referring to its census tract, as can be seen in table \ref{tabla-agentes}. When the simulation starts, each agent is randomly assigned to a location within their census tract. That location will be considered their ``home". In addition, depending on their economic activity, the agent will have another location to carry out  its profession. Thus, a worker agent will be linked with a work location, a student with a educational institution, and individuals on unemployment or retirement will not have any assigned locations. Each agent is also randomly assigned to one essential commerce location and one leisure location.

During this work, the term \textit{contact} will be used, not to refer to communication between agents, but to a physical contact between them that can lead to the transmission of the virus. This physical contact may occur beetwen any agent within the same location.

The model poses a simplified movement of agents on working weekdays and non-working weekdays. Any day of the week, all agents can have contact with their friends. Furthermore, on working weekdays, worker agents contact with other worker agents in the same place of work whereas non-worker agents will have contact with other non-worker agents in locations of essential commerce. Finally, on non-business days, agents contact with others at their current location, which can be both an essential or non-essential business location. Which days are considered business days or leisure days as well as the probability that an agent is in a location of leisure or essential commerce are fully configurable parameters through a configuration file.

\section{Graphical interface of the model}\label{sec-interfaz}

The model proposed in this work has been implemented using Netlogo, a  free-to-use integrated development environment (IDE) for ABM and multi-agent environments \cite{wilensky2015introduction}. Netlogo has a huge amount of freely accessible documentation and sample models. This has been the key reason for selecting Netlogo over other alternatives such as StarLogo TNG \cite{klopfer2009starlogo}, Mason \cite{luke2018mason} or Gama \cite{drogoul2013gama}.

\begin{figure*}
\centerline{\includegraphics[width=\textwidth]{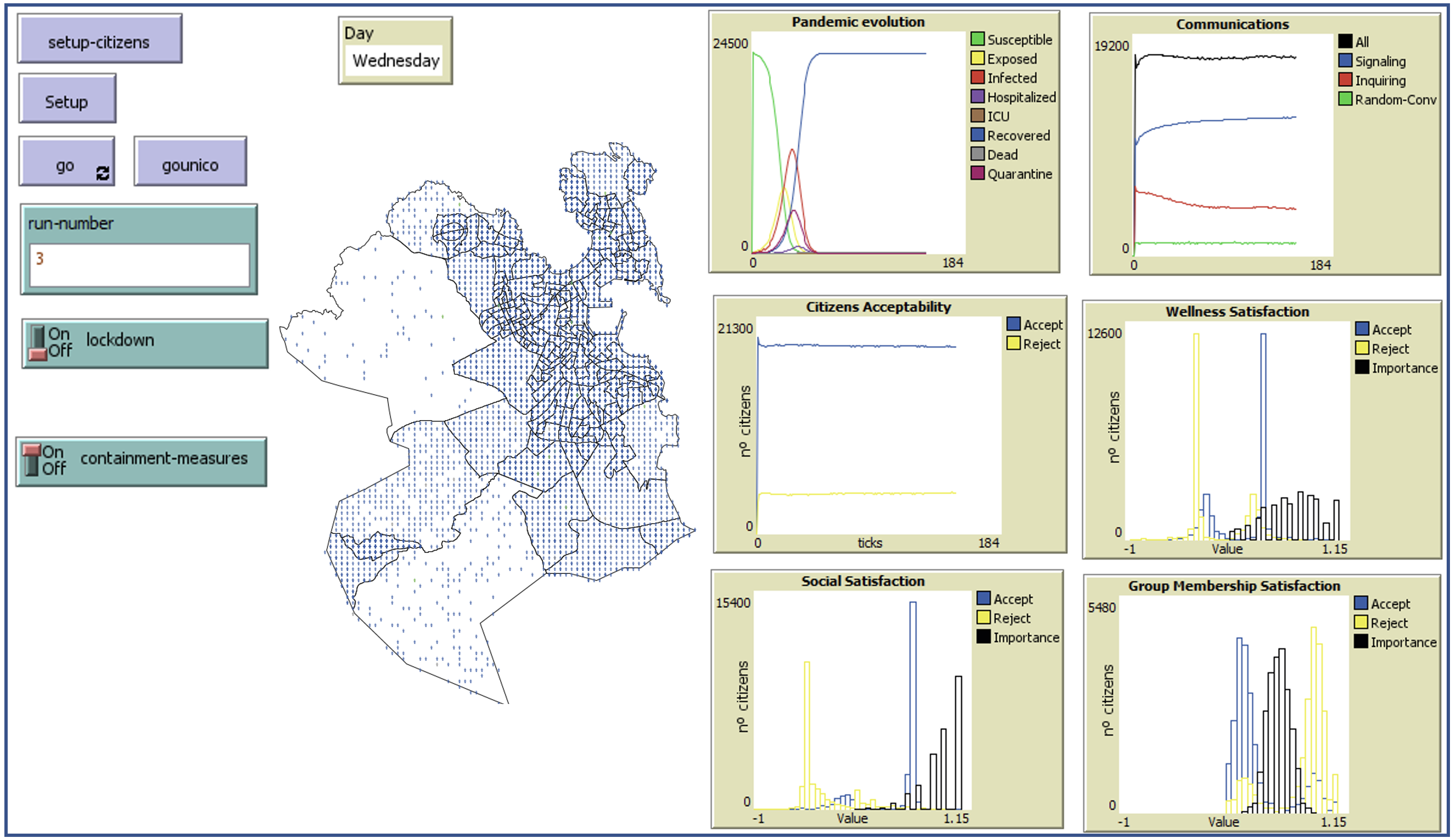}}
\caption{Model interface}
\label{fig-interfaz}
\end{figure*}

In figure \ref{fig-interfaz} we can see the final interface of the model. In the center, there is the 50x50 board representing the city with the agents located in their corresponding cells. There is also a small box showing the current day within the execution. On the left, we see the buttons for the initialization and execution of the simulation and selectors to switch between possible simulation scenarios (as it will be explained later). On the right, there are six graphs that, from left to right, represent: (1) the evolution of the pandemic (number of agents vs time in days) , (2) the number of communications between agents in the simulation, (3) the acceptability level of the measures; and three histograms with the \emph{importance} values of each of the needs as well as the \emph{satisfaction} generated by each of the agent's behavior alternatives (accept or reject). Each agent is colored to represent its current situation in the SEIRD transition diagram. Table \ref{tabla-estadoscolor} shows the state-color relationship.
 
 \begin{table}[htbp]
 \caption{Relationship between the colors of the agents in the interfaz and their state}
\begin{center}
\begin{tabular}{cc}
\toprule
\textbf{State} & \textbf{Color} \\
\midrule
Susceptible & Green\\
Exposed & Yellow\\
Infected & Red\\
Hospitalized & Purple \\
ICU & Brown \\
Quarantine & Magenta \\
Recovered & Blue\\
Dead & Gray \\
\bottomrule
\end{tabular}
\label{tabla-estadoscolor}
\end{center}
\end{table}

\section{Initialization and execution loop of the model}\label{sec-loop}

Running a model simulation has two distinct stages. First, the initialization of the virtual environment and the agents: board initialization, generation of the population, construction of social networks. 

\begin{figure*}
\centerline{\includegraphics[width = 10cm] {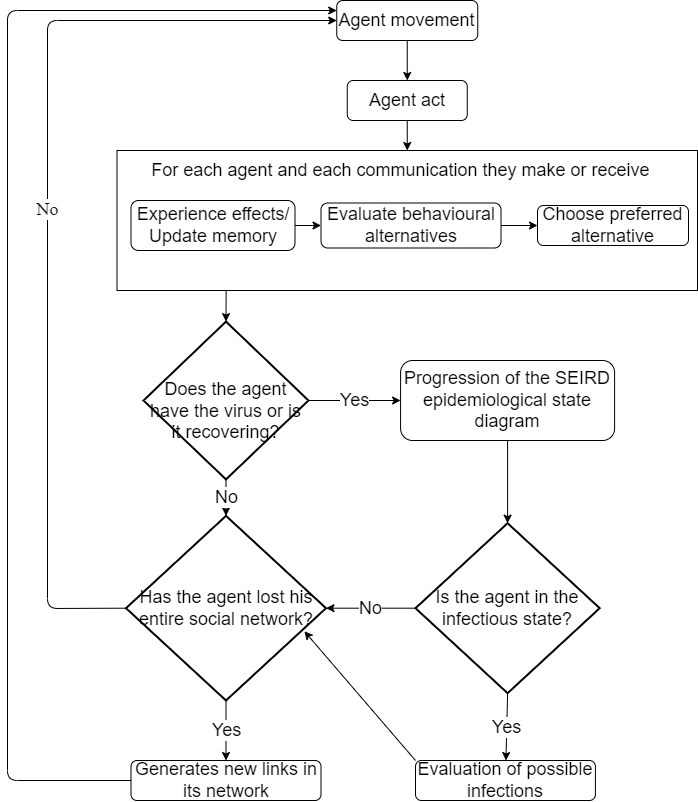}}
\caption{Model execution cycle}
\label{fig-core-loop}
\end{figure*}

Second, the main execution loop of the model. A graphical representation of the loop can be seen in figure \ref{fig-core-loop}. During each complete cycle of the model, which is equivalent to one day in the simulation, citizens: (1) move from their location to the appropriate work, leisure, or business location according to the day of the week and the economic activity of the agent himself, (2) they execute the HUMAT decision-making loop (see figure \ref{fig-humat}), (3) if they are COVID-19 patients, they evolve their status with respect to the virus following the figure \ref{fig-TransiciónEntreEstados}, (4) if they are in the infectious state they can infect other agents in their same location or in their network of friends and (5) if their social circle has disappeared due to the pandemic, they seek to relate to new agents by generating new links in the social networks of the model.

\section{Practical application: adaptation of the model to SARS-CoV-2 in the city of A Coruña}\label{sec-Ejemplo}

The proposed model aims to be used as a base architecture for epidemic simulations and thus being a tool to study the spread of a virus in a society and the effects of public reaction on the expansion.  To illustrate its use, we present an adaptation of the proposed general model to simulate the spread of the SARS-CoV-2 virus in the city of A Coruña (northern Spain).

\subsection{Adaptation of the Virtual Environment to A Coruña city.}

The first step for adapting the model to the case of SARS-CoV-2 in A Coruña is to represent the city on the two dimensional board. To do this, we start from the information provided by the National Institute of Statistics (INE) that contains the cartographic representation of Spain\footnote{ Available at \url{https://www.ine.es/censos2011_datos/cen11_datos_resultados_seccen.htm}. Last access 26 May 2022}.

In section \ref{sec-representacionsociedad} the mobility of agents between different locations has been described, for example, home-work or home-leisure travel. The locations are distributed in the environment thanks to a configuration file that indicates the geographical coordinates of each location, that is work, college, essential business, or non-essential business. Randomly, each agent is assigned to an essential business location, a non-essential business location, and, if applicable, a work or college location based on its business activity. These locations allow to create \emph{meeting points} where an agent is likely to be infected outside of their usual network of contacts, that is, their network of friends.

Once designed the geographic environment, the social network that links the agents to each other must be configured to finalize the virtual environment as described in section \ref{sub-entornovirtual}. 
To do this, one must set values to the parameters of the table \ref{tabla-socialnetworks}. 
For this adaptation we used the default values of circle radius, social reach and random friend probability. Remember that the links between the different agents represent the trust that one agent has in the other and, therefore, it is also necessary to establish this value. These links are bidirectional (if agent A is a friend of agent B, B should be a friend of A), but not reciprocal (the trust that A has in B does not have to be equal to the trust that B has in A). Given that there is no information available about the trust that each agent has in the individuals pertaining to his social network, each link has been assigned a value that follows a uniform distribution in the range $[0,1]$.

Once the environment has been configured, the population must be generated, that is, the agents. Next subsection details how these agents have been created.

\subsection{Population of A Coruña}\label{sec-cityPopulation}

The presented model operates at the level of citizens, so personal attribute information is needed for the entire population of A Coruña to calibrate, validate and apply the model. This information will allow us to test different scenarios of virus containment measures. Table \ref{tabla-agentes} indicates the demographic variables as well as the satisfaction and importance of each need required for each agent. All these data were acquired for a certain number of citizens of A Coruña through a survey. These citizens will be directly represented by agents in the system. However, as the population is larger we have developed a method to complete the model with more \emph{simulated} agents representing the city, as described below.

\paragraph{Generation of \textit{real} agents}. 
A survey was considered as an adequate way to collect data from citizens. This survey measures both the situational characteristics (place of residence, type of housing, etc.) and the psychological characteristics of the population in the midst of the generated pandemic by SARS-CoV-2.  A psychologist supported its design and it was disseminated through electronic media and social networks in May 2020.

The survey included questions about the position, for or against, on the preventive measures proposed by the government in relation to SARS-CoV-2, questions about sociodemographic data (see table \ref{survey-table}), as well as about psychological characteristics of the population or their life situation. The psychological questions were grouped according to the need it refers to and whether it refers to the importance given to that need or to its satisfaction. Finally, an average was made for each need to determine its importance and its satisfaction, except for the need to belong to the group, which is dependent on the individual's social network \ref{sec-Agentes}.

Finally, we collected 1,274 valid surveys whose data were used to create the initial population of the city of A Coruña. However, the number of inhabitants in A Coruña on January 1st, 2021, is 245,468 \cite{IneHabitantes}, so using only 1,275 agents in the model would mean using 0.5\% of the inhabitants, carrying possible problems in the calibration of the model and in the representation of its reality. Therefore, \textit{simulated} agents must be generated to increase the population of the model. Next subsections will explain how to achieve them.

\begin{table*}
  \caption{Sociodemographic variables of the survey}
  \label{survey-table}
  \begin{tabular}{m{0.3\textwidth}m{0.6\textwidth}}
    \toprule
    \textbf{Parameter} & \textbf{Value} \\
    \midrule
    gender & man/woman\\
    age & integer in [18,100]\\
    family & one-person household/ father or mother alone with children / father or mother alone with children and other people/ Couple with children/ couple with children and other people/ Couple without children / Another type of home \\ 
    children under 12 years old &  yes/no \\ 
    rural house & yes/no\\
    dwelling with garden & yes/no\\
    dwelling size & square meters\\
    economic activity & employee/ unemployed/ autonomous/ civil servant / executive or director/ college student / retired\\
    essential work & yes/no\\
    net monthly salary in euros & no income / less than 1000 / [1000-1500]/  [1501-3000]/  [3001-4500]/  [4501-6000]/ more than 6000 \\
    census tract & census tract code\\
    \bottomrule
  \end{tabular}
\end{table*}


\vspace{0.2cm}
\noindent \textbf{Generation of citizen profiles}. To generate the artificial population, following the process described in \cite{alonso2021generating}, citizen profiles were created that relate sociodemographic and psychological data to each individual's decision on whether or not to comply with the proposed preventive measures. To find this relation decision trees were used, since the solution they provide is easy to explain and to be validated by our psychologist. To train this decision tree,  based on the survey carried out, we employed as inputs the variables: gender, age, family, children under 12 years of age, area of the dwelling, type of home, square meters of the dwelling (reduced to a large or small house), economic activity, essential work, and net income. Those variables with multiple classes, such as family and economic activity, were transformed into multiple binary variables following a \textit{one-hot encoder} scheme. The type of essential work was discarded due to a large number of response values (more than 60).

Regarding to the desired outcome that the decision tree should obtain, the survey included two questions directly related: (1) is respecting the prevention measures an act of solidarity? and, (2) does respecting the measures protect me from contagion?. The survey asks these questions on a Likert scale from 1 to 5, thus for training the decision tree, this  scale was converted to three possible answers: No (Likert values 1 and 2), Undecided (3), and Yes (4 and 5) leading to the classes, as shown in table \ref{tabla-datos-salida-arbol}, in which our citizens can be classified.

\begin{table*}
\caption{Desired outcomes for the decision tree retrieved from the survey as the main reasons to support the preventive measures.}
\begin{center}
\begin{tabular}{cccc}
\toprule
\textbf{Question} & \textbf{Yes} & \textbf{Undecided} & \textbf{No}\\
\midrule
Act of solidarity & 1143 & 67 & 69\\
Protection & 1142 & 97 & 36\\
\bottomrule
\end{tabular}
\label{tabla-datos-salida-arbol}
\end{center}
\end{table*}

As can be seen in table \ref{tabla-datos-salida-arbol} the sample is highly unbalanced, with a high percentage of the population (90\% in the worst case) supporting the measures.

To balance the data, the sampling reduction method \textit{random under-sampling (RUS)} \cite{devi2020review} was applied to the majority class and the oversampling technique \textit{Synthetic Minority Over-sampling Technique (SMOTE)} \cite{chawla2002smote} was applied to the underrepresented classes. Finally, the minority classes \textit{Undecided} and \textit{No} were grouped into a single class \textit{Reject} for the purpose of further balancing the set.

Different decision tree configurations were trained for each one of both questions, changing parameters such as the minimum number of samples allowed in a leaf node or the maximum number of tree splits. Finally, the tree with the best performance was obtained using the impurity criterion as a measure to generate a division and impose a minimum value of 35 samples per leaf.

\begin{figure*}
\centerline{\includegraphics[width= 10cm]{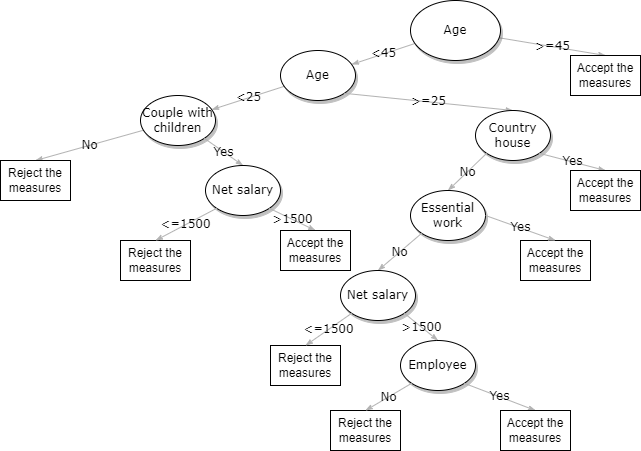}}
\caption{Decision tree to classify citizens according to the question: Is respecting prevention measures an act of solidarity?}
\label{fig-tree-solidarity}
\end{figure*}

For the question, \emph{Does respecting the measures protect me from contagion?}, a tree was obtained with an overall precision of $0.71$. For the question, \emph{Is respecting the prevention measures an act of solidarity?}, the global precision obtained was $0.76$. Finally, the latter was selected for having the best overall accuracy. The resulting tree is shown in Figure \ref{fig-tree-solidarity}.

Notice that each branch that leads to a leaf node of the tree defines a citizen profile. For example, if we follow the left branch of the tree in the figure \ref{fig-tree-solidarity}, we will see that a profile that rejects the measures is that formed by people under 24 years of age who live with their partner and do not have children.

From this decision tree, we concluded that the most relevant variables that determine the profiles of the people who accept or reject the measures are: age, type of family unit, net income, whether or not they live in a rural house, whether or not they carry out an essential job and whether or not they are in a paid employment. Based on the profiles determined by the tree, more agents will be created to complete the population, as explained below.

\vspace{0.2cm}
\noindent \textbf{Generation of \textit{simulated} agents}. The simulated agents were generated using three different the census data of A Coruña, the survey, and the previously obtained decision tree, as follows \cite{alonso2021generating}:

\begin{enumerate}
    \item Census data provides information on the age and gender distribution for each census tract in the city. Therefore, agents were generated to maintain these proportions for each census section.
    \item The survey data was used to complete the rest of the sociodemographic variables (see table \ref{tabla-agentes}) for these new agents. First, for each variable determined as relevant by the decision tree, its distribution in the survey was calculated. To obtain these distributions, the survey data is divided according to gender and the age clusters determined by the tree. For example, to obtain the distribution of the "net salary" variable in the survey, this distribution is calculated in different age and gender clusters, such as women between 25 and 50 years old or men between 18 and 25 years old. Following these distributions the sociodemographic variables of the simulated agents are filled.
    \item Up to this point the simulated agents have values for all the variables except for those related to their needs (satisfaction and importance). Subsequently, these variables are taken from those of the \textit{real} agents. To do this, each \textit{simulated} agent is classified, according to its sociodemographic variables, in one or another leaf (profile) of the tree. Once classified, their behavior regarding their needs  is taken from a \textit{real} agent randomly chosen among those belonging to the same profile. It should be noted that the leaf nodes are not pure and within the same citizen profile there are people who approve the measures as well as people who reject them. These proportions are also taken into account.

\end{enumerate}

\subsection{Adaptation of SEIRD to SARS-CoV-2}

The first question to answer before using the basic model proposed in this work is to verify that the virus to be modeled complies with the SEIRD state transition (see figure \ref{fig-TransiciónEntreEstados}) or if it would be easily adaptable. In the specific case of SARS-CoV-2, the SEIRD diagram fits perfectly with the evolution of the virus. A healthy individual becomes infected, goes through an incubation period until it reaches an infectious stage, where it can infect others with or without symptoms, and finally dies or recovers. In addition, when an individual recovers from the virus, he is resistant to it for a time but eventually returns to the initial state where he can be infected again. In addition, the patient goes through the added states in the infectious state, concerning hospitalization and admission to the intensive care unit by the patient.

Once verified that the stages of virus transmission fit into the SEIRD scheme, the variables in figure \ref{fig-TransiciónEntreEstados} must be adapted to the virus. Currently, the pandemic caused by SARS-CoV-2 is still ongoing, which means that both the reference data and the virus itself, and our knowledge about it, are constantly changing and evolving. For this reason, it is difficult to obtain specific data on the rate of contagion or the duration of the disease in patients. In table \ref{tabla-covid-19} we can see each of the values chosen for the adaptation of the transition diagram to SARS-CoV-2, citing the scientific-technical publication from which it was extracted. Those values without a reference will need future calibration so that the model faithfully recreates the desired situation.

\begin{table}[htbp]
\caption{SARS-CoV-2 transition data}
\begin{center}
\begin{tabular}{cc}
\toprule
\textbf{Parameter} & \textbf{Value}\\
\midrule
P-SE & 0.07\\
P-ID & 0.005 \cite{informe-renave0309}\\
P-IH & 0.07 \cite{informe-renave0309}\\
P-IR & 1 - (P-id + P-ih)\\
P-HD & 0.005 \cite{informe-renave0309}\\
P-HICU & 0.08 \cite{informe-renave0309}\\
P-HR & 1 - (P-HD + P-HICU)\\
P-ICUD & 0.31 \cite{ciberes}\\
P-ICUR & 1 - P-ICUD\\
days-EI & Lognormal $\mu$ = 1.621, $\sigma$ = 0.418 \cite{lauer2020incubation}\\
days-IH & 5\\
days-ID & 10 \cite{ECDC}\\
days-IR & 10 \cite{ECDC}\\
days-HICU & 3\\
days-HD & 10 \cite{ECDC}\\
days-HR & 10 \cite{ECDC}\\
days-ICUD & 7\\
days-ICUR & 7\\
\bottomrule
\end{tabular}
\label{tabla-covid-19}
\end{center}
\end{table}

\subsection{Model Temporality}

The model execution, simulating the expansion of SARS-COV-2 in the city of A Coruña, cover approximately the period between the beginning of August 2020 and the end of the same year. These dates coincide, respectively, with both the second wave in A Coruña and the starting of the vaccination in Spain \cite{SERGAS}. Therefore, the model simulates 150 days where each actual day equals one cycle in the model run.

To determine which of these days are working days or non-working days, the most widely used criterion has been used. Thus, Monday through Friday are considered working days and workers or students will go to their job or class, while Saturday and Sunday will be non-working days and citizen agents will be distributed between essential and non-essential points of commerce. The probability that an agent visits a non-essential place of business is defined by a configurable parameter with a default value of 0.25. 


\section{Evaluation of alternative scenarios}\label{sec-Escenarios}

We propose three different scenarios to demonstrate the adaptability of the model to the different degrees of virus containment measures: (1) a lax scenario without any measure, (2) a very restrictive scenario with total confinement, and (3) a more realistic scenario that considers some preventive measures that governments applied. The number of simulated agents is 11,646, representing approximately 5\% of the population of A Coruña. The proportion of infected agents was the one at the beginning of the second wave in A Coruña (August 2020).

\subsection{First scenario: no measures} 
In the first scenario, the simulations are carried out without taking into account any preventive or containment measures using the SEIRD model adapted for SARS-CoV-2 as already described. In this scenario, the critical nodes are not taken into account since, as there are no preventive measures, it is not possible to influence the citizens on them.

\begin{figure}
\centering
\begin{minipage}[b]{0.5\textwidth}
    \centering
    \includegraphics[width=.95\linewidth]{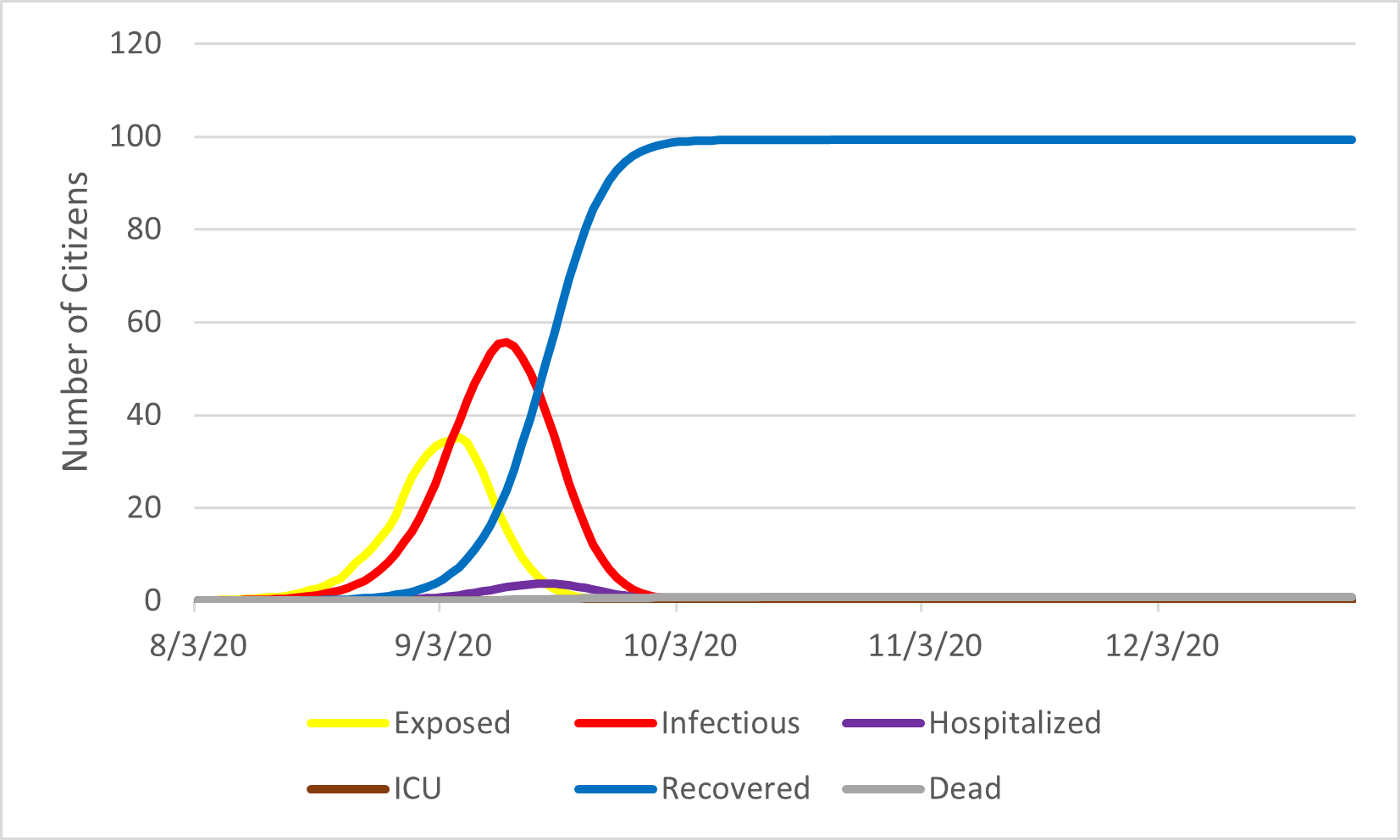}  
    \caption{Evolution of the epidemic in the 1st scenario}
    \label{fig-escenario1}
    \centering
    \includegraphics[width=.95\linewidth]{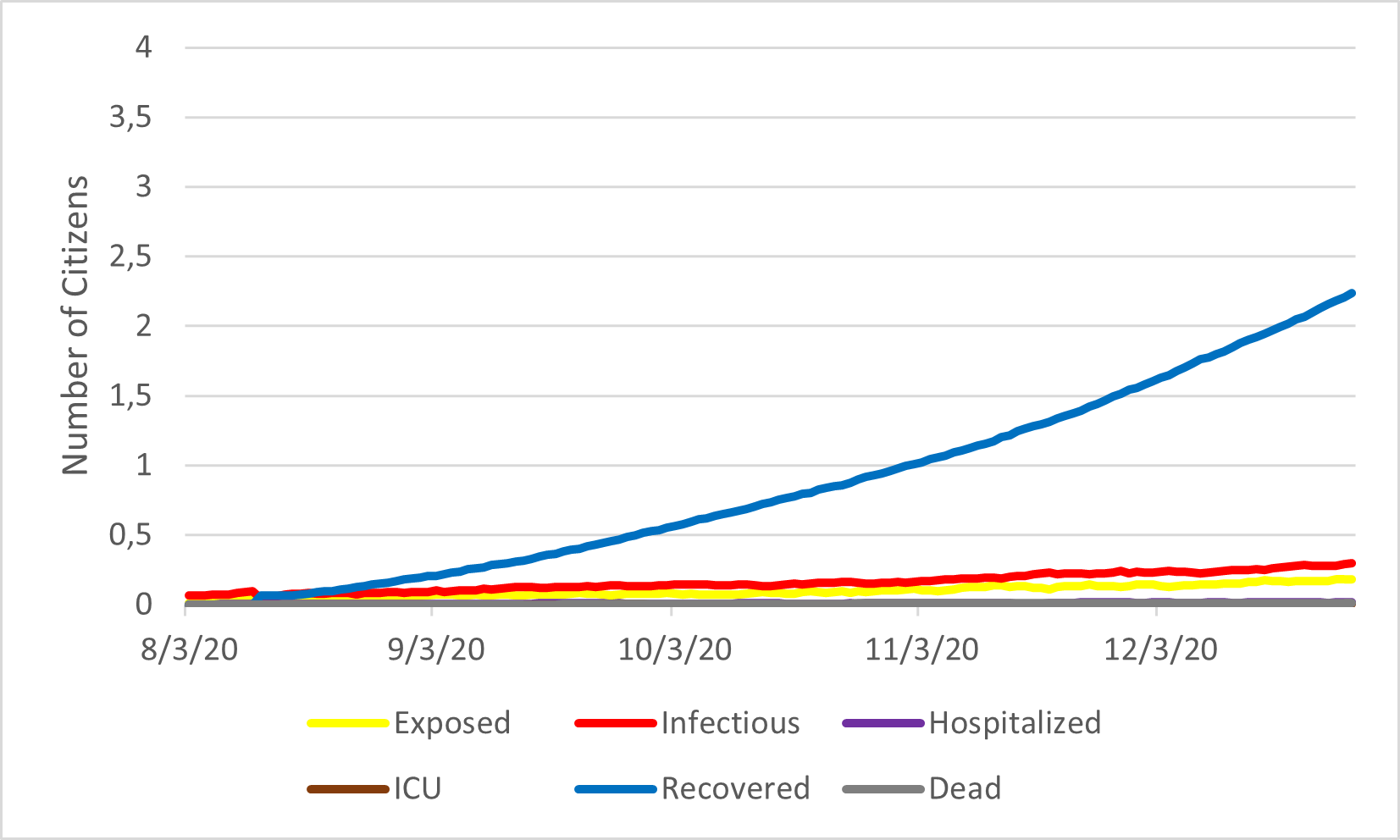}  
    \caption{Evolution of the epidemic in the 2nd scenario}
    \label{fig-escenario2}
\end{minipage}%
\begin{minipage}[b]{0.5\textwidth}
    \centering
    \includegraphics[width=.95\linewidth]{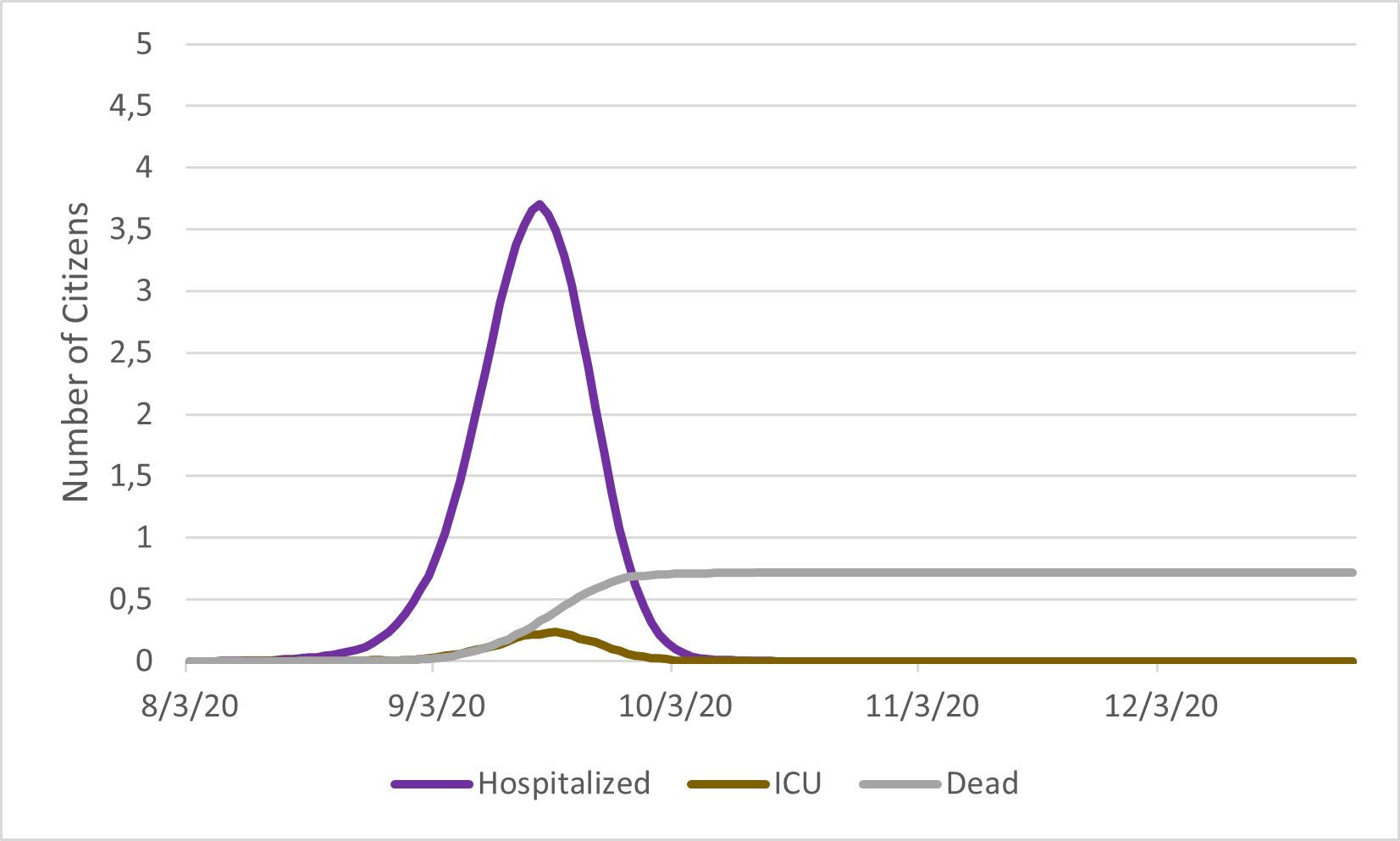}  
    \caption{Evolution of hospitalized patients in the 1st scenario}
    \label{fig-escenario1Zoom}
    \centering
    \includegraphics[width=.95\linewidth]{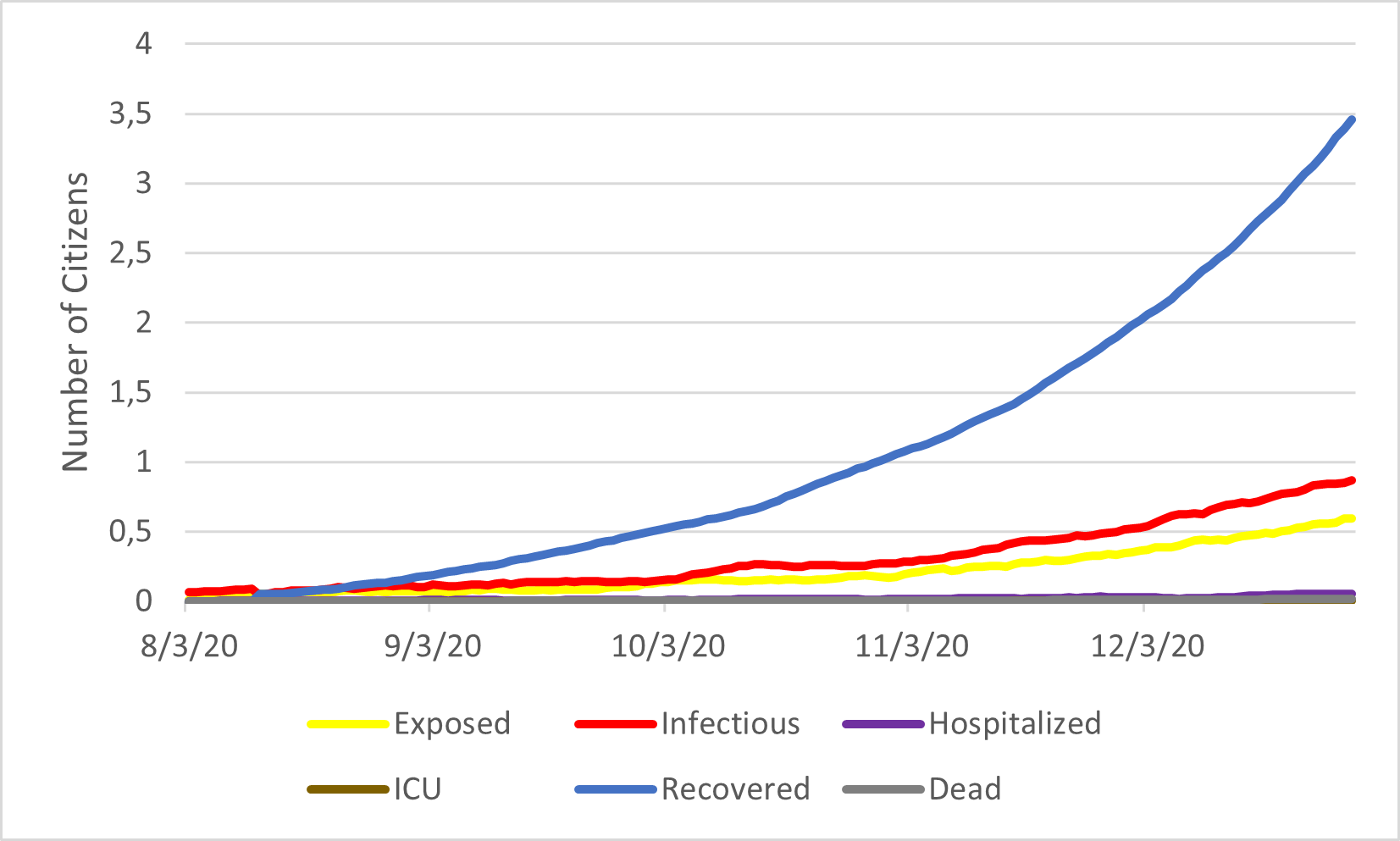}  
    \caption{Evolution of the epidemic in the 3rd scenario}
    \label{fig-escenario3}
\end{minipage}
\caption{Evolution of the epidemic in the proposed alternative scenarios. The first scenario (a) is a scenario free of preventive measures; hospital inflows and deaths in this scenario are represented in a new graph (b) to facilitate readability. The second example (c) recreates a very restrictive situation. Finally, scenario 3(d) exemplifies an intermediate situation where some preventive measures are in place.}
\label{fig-escenarios}
\end{figure}

In figure \ref{fig-escenario1} we can see the average results of the simulation of the model for 10 executions. The graph reflects the number of agents in each state of the SEIRD model for each day of the period the model was run: number of infected (red), exposed (yellow), hospitalized (purple), ICU (brown), recovered (blue), and dead (gray).  The evolution of hospitalized, in the ICU, and dead citizens is shown in more detail in figure \ref{fig-escenario1Zoom}. We can observe, from graph \ref{fig-escenario1}, that the entire population is infected in a couple of months if no restrictive measure is applied to control the epidemic: in mid-September, a peak of more than 6,000 infected inhabitants (around 50\% of the population) is reached and, after two months, practically all the agents have recovered from the disease. Also, in graph \ref{fig-escenario1Zoom} it can be observed that deaths and serious hospitalizations increase during the peak of the pandemic, as expected.

\subsection{Second scenario: quarantine} 
For the second scenario, the simulation interval is the same as in the previous scenario but declares a state of complete confinement. Accordingly, the behavior of the agents will be modified as follows:

\begin{itemize}
    \item Only citizens who have an essential job will go to the work location.
    \item Non-essential commerce locations will remain closed.
    \item Compliance with security measures at essential business locations will be mandatory.
    \item Agents will be in quarantine, so they will have no chance to spread the virus to their network of friends. Since communications are maintained, they continue to influence each other.
    \item Citizens that decide not to comply with the measures will hold meetings with the members of their network of friends who also do not comply with the measures.
    \item The probability of contagion is reduced compared to the scenario without measures using a P-SE of 0.05.
\end{itemize}

In this scenario, the critical nodes could be used, but in Spain the stages of total confinement were strongly monitored by the state security forces, making massive non-compliance impossible.

In figure \ref{fig-escenario2} we can see the average results of 10 simulations for this scenario. The evolution of the infected graph is in lower values than in the previous scenario, demonstrating the effectiveness of more restrictive measures. If we look at the red line, representing the number of infected, we can see that the growth is gradual and practically non-existent compared to scenario 1.

\subsection{Third scenario: Application of Preventive Measures} 

 At this stage, preventive measures are those imposed by the Government of Spain when the strict lockdown ended and include the mandatory use of a mask on the street and in closed rooms, a safety distance of 1.5 meters, and capacity control inside premises \cite{gobiernoespana}. In addition, individual confinement is contemplated in case of testing positive in an antigen or polymerase chain reaction test (PCR) or being in contact with a positive one recently.

To represent these measures in the model it is necessary to apply some changes to the epidemiological transition diagram:

\begin{enumerate}
    \item First, to reflect the probability of the transition between the states \textit{Susceptible} and \textit{Infected} in this scenario we used three different probabilities:
    \begin{itemize}
        \item P-SE. The probability of contagion of an individual who does not respect prevention measures.
        \item P-SE-no-essential-business. The probability of contagion of an individual who respects prevention measures at a non-essential business location.
        \item P-SE-accepting. The probability of contagion of an individual who respects prevention measures at a essential business location.
    \end{itemize}
    The reason to consider these last two probabilities separately is because, in general, in essential businesses the requirement to comply with the measures is greater and, therefore, the probability of contamination is lower.
    \item Second, a new substate of the \textit{Infectious} state called \textit{Quarantine} is added, as an agent in this state cannot infect others. In this scenario, an agent will enter quarantine 3 days after entering the infectious state, and will leave quarantine when they recover or decide not to follow prevention measures. As an exception, some agents are asymptomatic, and they are not in quarantine because they are not aware of being infected. For this adaptation, the percentage of asymptomatic patients is 40\% \cite{ma2021global}. 
\end{enumerate}

In graph \ref{fig-escenario3} we can see the average results of the simulation of the model for 10 executions. If this graph is compared with the graph \ref{fig-escenario2} of the second scenario, we can see that the preventive measures work slightly worse than total confinement, but if it is compared with the graph \ref{fig-escenario1} of the first scenario, it can be concluded that the preventive measures are effective in containing the virus.


The third scenario is the best example to study the effect of critical nodes on the evolution of the epidemic. Preventive measures are individual responsibility, so the level of acceptance of each citizen affects compliance with them. Based on this scenario, two new ones are designed to test the impact of critical nodes on the behavior and opinion of citizens, including the consequences of these changes on the evolution of the epidemic.

In the first alternative, an utopia is assumed where all critical nodes help to alleviate the pandemic's impact on society. With this aim, two critical nodes are included: the city council and the press. Every five days the City Council will send communications to the public in favor of preventive measures through the press.

In the second alternative, a more realistic situation arises where we have two main critical nodes: the city council that launches communications in favor of the measures, and the political opposition that intends to influence against the use of preventive measures \footnote{The political opposition raises doubts about the proposed measures, alleging their ineffectiveness compared to alternatives that it would use}. Again, both nodes send their communications every 5 days through press.

\begin{figure}
\centering
\begin{minipage}[b]{0.5\textwidth}
    \includegraphics[width=.95\linewidth]{Figures/Escenario3-Colored-Percentage.png}
    \caption{Evolution of the epidemic in scenario 3}
    \label{fig-escenario3-comp}
    \centering
    \includegraphics[width=.95\linewidth]{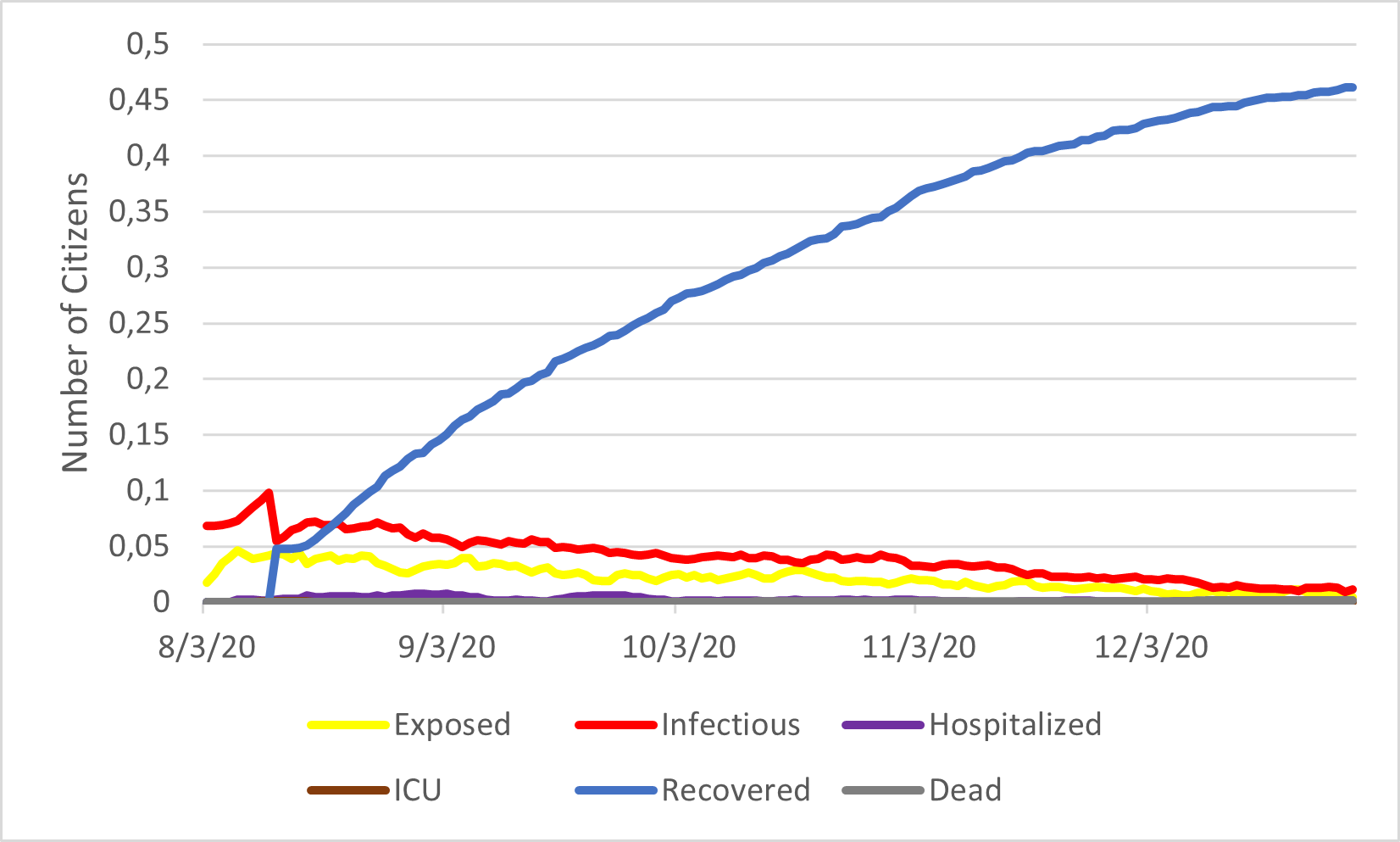}  
    \caption{Evolution of the epidemic in scenario 3A}
    \label{fig-escenario3A}
\end{minipage}%
\begin{minipage}[b]{0.5\textwidth}
    \includegraphics[width=.95\linewidth]{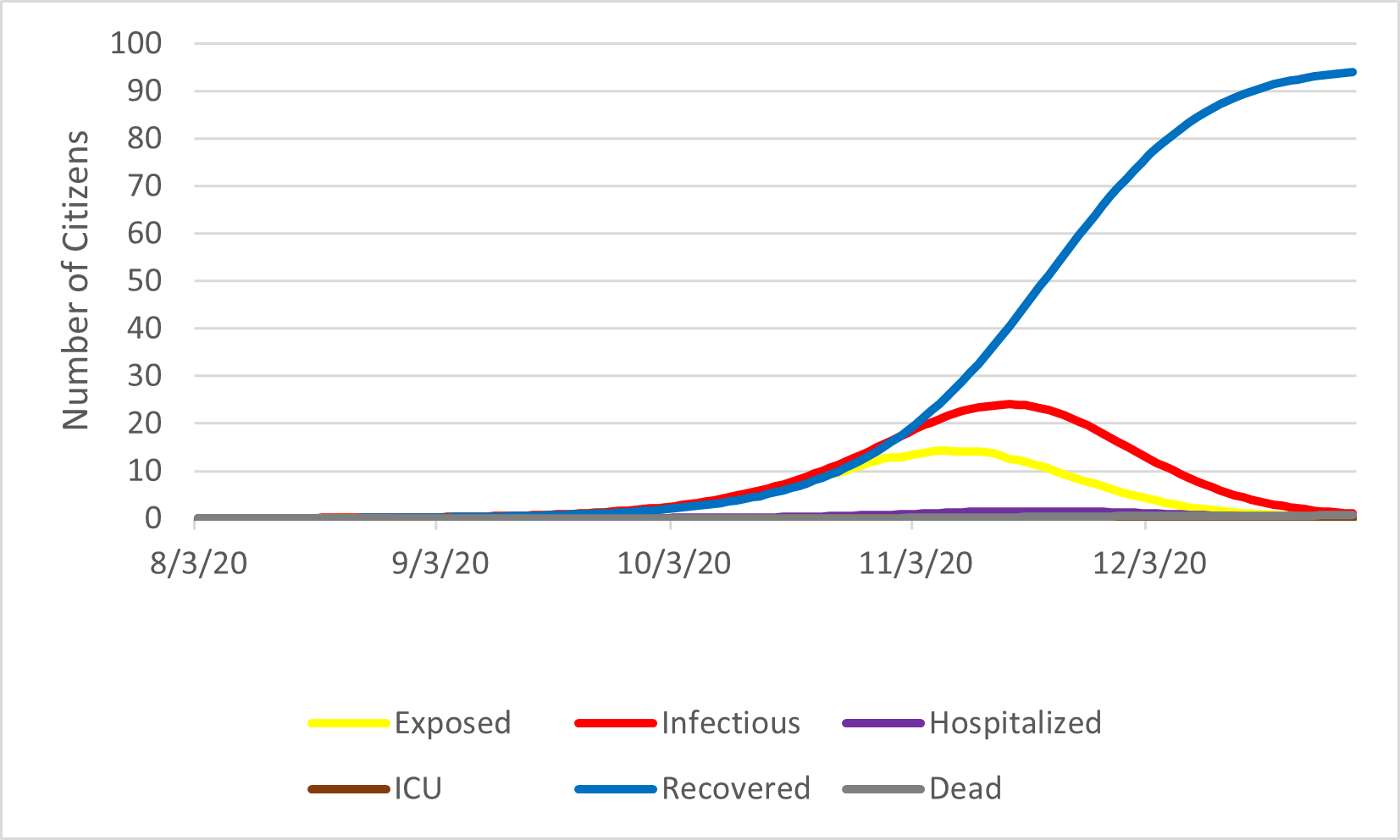}  
    \caption{Evolution of hospitalized patients in scenario 3B}
    \label{fig-escenario3B}
\end{minipage}
\caption{Evolution of the epidemic: (a) scenario 3, (b) variation in which there is only influence in favor of the measures, (c) variation in which the critical nodes influence for and against the measures.}
\label{fig-escenario3-variations}
\end{figure}

In the figure \ref{fig-escenario3-variations}, we can see a comparison between the original scenario 3 and its alternatives with critical nodes in operation. Scenario 3A is where the critical nodes only influence in favor of the measures, causing greater adherence by the community, which results in greater containment of the virus. However, in scenario 3B there are nodes for and against the measures. That causes a drop in acceptability and a lack of control in the evolution of the pandemic.

With these three scenarios, it has been shown that the model is capable of adapting to simulations with different prevention measures. The first example proposes a scenario free of preventive measures where the evolution of infections increases rapidly, whereas the second example recreates a very restrictive situation where infections grow very slowly and very few citizens end up suffering from the disease.  Finally, scenario 3 exemplifies an intermediate situation where there are some prevention measures, but citizens lead a relatively normal life. In this simulation, infections are higher than in confinement, but they are far from the situation in scenario 1.

\section{Conclusions and future work}
Classic epidemiological models allow studying the spread of a virus in a society, but ignore the specific characteristics of the individuals that make it up. Agent-based models can be a solution since they allow individuals to be represented based on their own characteristics, permitting the study of the impact of their actions on the global spread of contagion. In addition, there are socio-psychological issues and circumstances of the individual's environment, normally not included in the classic models of expansion, that influence this attitude. The agent-based model that has been presented proposes a general architecture that allows to consider these issues that have proven to be of great importance in pandemic situations such as that of COVID-19, in which the attitude of citizens as well as interpersonal trust and government credibility has proven to be decisive in containing the virus. In summary, the ABM architecture has been proposed that is easily adaptable to any population or virus and whose main characteristics are: 
\begin{itemize}
    \item It uses an adaptation of the SEIRD model to simulate the spread of the virus.
    \item It uses the HUMAT decision-making model to evaluate the possible behavior alternatives of citizens against the preventive measures proposed in each possible simulation scenario.
    \item It establishes two social networks (friendship and neighborhood) that will influence the decision-making of each citizen and determine links between them that allow the spread of the virus.
    \item It evaluates the effect of communications made by groups or institutions on the behavior of citizens and the effect that this change of behavior has on the epidemiological evolution.
    \item It places citizens in a 2D virtual environment that represents the city or region affected by the virus.
    \item It considers mobility of citizens (between home and work/study or leisure areas) which, through a relationship of proximity, can favor the spread of the virus.
\end{itemize}

The operation of the model and its possibilities of adaptation have been illustrated with the case of the expansion of SARS-CoV-2 in the city of A Coruña. The execution of the chosen scenarios, which range from strict confinement to not applying any preventive measure, denotes the adequacy of the model to represent the behavior of a population and its reactions. Therefore, it is a useful tool to test different containment policies and thus find the best balance between the harshness and effectiveness of the measure, partly determined by the adherence that it is capable of generating in the population. However, its complexity leaves us open lines for future work. Among all of them, it is worth mentioning:
\begin{itemize}
\item The adaptation of the SEIRD model involves a large number of parameters (see Table \ref{tabla-covid-19}). In addition, many of the values established for some of these parameters are open to question, since the information available on the behavior of SARS-CoV-2 (with its respective variants) has not stopped being updated day by day. For all these reasons, carrying out an adequate calibration, and exhaustively analyzing these parameters, is a very complex and time-consuming task (around thousands of hours as simulation of the A Coruña model lasts approximately 15 hours). 

\item The mobility of citizens can determine the spread of some viruses, especially those that spread by air and not by physical contact. For the lack of simplicity, the locations of leisure, work, or home have been established randomly and, likewise, the movements made by the agents between them. It would be desirable to set these locations more precisely and make allocations based on proximity in some cases (for example, essential business).
\item The agents are the key part of this model and, despite the success of the sample collected trough the survey, it would be desirable to obtain more data in order to have a population that is as realistic as possible. In addition, this would allow the survey to be expanded, including more specific questions and in accordance with the needs of the model that were not initially considered.
\end{itemize}

\begin{acks}
This work has been supported by GAIN (Galician Innovation Agency) and the Regional Ministry of Economy, Employment and Industry, Xunta de Galicia grant COV20/00604, through the ERDF, and grant ED431C 2022/44. We also wish to thank Dr.Adina Dumitru who helped to configure and evaluate the survey. 
\end{acks}

\appendix
\bibliographystyle{ACM-Reference-Format}
\bibliography{main}

\end{document}